%% file: main.tex
\pdfoutput=1
\documentclass[11pt]{article}

\usepackage{acl}
\usepackage{graphicx}
\usepackage{times}
\usepackage{latexsym}
\usepackage{tabularx}
\usepackage[T1]{fontenc}
\usepackage[utf8]{inputenc}

\usepackage{microtype}
\usepackage{booktabs}
\usepackage{xcolor}

\title{SemEval 2023 Task 6: LegalEval - Understanding Legal Texts}

\author{{\bf Ashutosh Modi}$^\mathparagraph$ \qquad
Prathamesh Kalamkar$^\dagger$ \qquad Saurabh Karn$^\ddagger$ \\ {\bf Aman Tiwari}$^\dagger$ \qquad {\bf Abhinav Joshi}$^\mathparagraph$\ \qquad {\bf Sai Kiran Tanikella}$^\mathparagraph$ \qquad \\  {\bf Shouvik Kumar Guha}$^\diamond$ \qquad {\bf Sachin Malhan}$^\ddagger$ \qquad {\bf Vivek Raghavan}$^\star$ 
 \\ 
        $^\mathparagraph$Indian Institute of Technology Kanpur (IIT-K), \\
        $^\dagger$Thoughtworks,
        $^\ddagger$Agami,
        $^\star$EkStep, $^\diamond$NUJS, West Bengal\\
  \texttt{\{ashutoshm, ajoshi, tskiran\}@cse.iitk.ac.in}, \\  
  \texttt{\{prathamk, aman.tiwari\}@thoughtworks.com}, \\
  \texttt{\{saurabh, sachin\}@agami.in}, \\
  \texttt{shouvikkumarguha@nujs.edu}, \texttt{vivek@ekstep.org} 
}

\begin{document}
\maketitle

\input{abstract}
\vspace{-2mm}
\input{introduction}

\vspace{-2mm}
\input{relatedwork}
\vspace{-5mm}
\input{tasks}
\vspace{-2mm}
\input{corpus_annotations}
\vspace{-2mm}
\input{teams}
\input{results}
\input{conclusion}

\section*{Acknowledgements}

We would like to thank legal experts who contributed in annotating datasets. In particular, we thank Shreyangshi Gupta, Priyanka Choudhary, Srinjoy Das and Khushi Jain for annotating CJPE test set. 

\bibliography{custom}

%\appendix

\end{document}

%% file: abstract.tex
\begin{abstract}
In populous countries, pending legal cases have been growing exponentially. There is a need for developing NLP-based techniques for processing and automatically understanding legal documents. To promote research in the area of Legal NLP we organized the shared task LegalEval - Understanding Legal Texts  at SemEval 2023. LegalEval task has three sub-tasks: Task-A (Rhetorical Roles Labeling) is about automatically structuring legal documents into semantically coherent units, Task-B (Legal Named Entity Recognition) deals with identifying relevant entities in a legal document and Task-C (Court Judgement Prediction with Explanation) explores the possibility of automatically predicting the outcome of a legal case along with providing an explanation for the prediction. In total 26 teams (approx. 100 participants spread across the world) submitted systems paper. In each of the sub-tasks, the proposed systems outperformed the baselines; however, there is a lot of scope for improvement. This paper describes the tasks, and analyzes techniques proposed by various teams. % for each of the sub-tasks and analyzes the models and results of each of the teams.   

%We address these concerns by proposing three shared tasks: (1) structuring unstructured legal documents into semantically coherent units (Rhetorical Roles), (2) identifying relevant entities in a legal document (Named Entity Recognition), and (3) predicting the outcome of a case along with an explanation (Court Judgement Prediction with Explanation). For each task, we have created annotated datasets and baseline models.    

%Automatic Structuring of Court Judgments using sentence rhetorical roles. In particular, we introduce a corpus of legal judgment documents in English that are segmented into topical and coherent parts. Each of these parts is annotated with a label coming from a list of pre-defined Rhetorical Roles. This task can be treated as sequential sentence classification task. 
\end{abstract}

%% file: introduction.tex
\section{Introduction} \label{sec:intro}

In populous countries (e.g., India), pending legal cases have grown exponentially. Due to the nature of the legal domain, it may not be possible to automate the entire judicial pipeline completely; nevertheless, many intermediate tasks can be automated to augment legal practitioners and hence expedite the system. For example, legal documents can be processed with the help of Natural Language Processing (NLP) techniques to organize and structure the data to be amenable to automatic search and retrieval. However, legal texts are different from commonly occurring texts typically used to train NLP models. Legal documents are quite long (tens and sometimes hundreds of pages), making it challenging to process \cite{malik-etal-2021-ildc}. Another challenge with legal documents is that they use different lexicons compared to standard texts. Moreover, legal documents are manually typed in countries like India and are highly unstructured and noisy (e.g., spelling and grammatical mistakes). Above mentioned challenges make it difficult to apply existing NLP models and techniques directly, which calls for the development of legal domain-specific techniques.

To promote research and development in the area of Legal-AI we proposed shared tasks via ``LegalEval - Understanding Legal Texts". This paper gives details about the shared task and our experience in organizing LegalEval. In total 26 teams submitted systems papers in the shared task. Though the documents pertained to the Indian legal system, teams across the globe participated showing an interest in developing legal-NLP techniques that could later be generalized to other legal systems as well. 

The paper is organized as follows. In Section \ref{sec:related-work}, we outline work done in the legal-NLP domain. Section \ref{sec:tasks} describes the tasks in detail. This is followed by details of the corpora (used for various sub-tasks) in Section \ref{sec:corpus}. Details of models developed by each of the teams are discussed in Section \ref{sec:teams}. Section \ref{sec:results} analyzes the results furnished by various techniques for each of the sub-task. Finally, Section \ref{sec:conclusion} summarizes and concludes the paper.

%Though legal documents use natural language (e.g., English), many commonly occurring words/terms have different legal connotations. The use of different lexicons makes it challenging to adapt existing NLP models to legal texts \cite{malik-etal-2021-ildc}.

%% file: relatedwork.tex
\section{Related Work} \label{sec:related-work}

In recent years, Legal NLP has become an active area for research. Researchers have worked on various research problems pertaining to the legal domain such as Legal Judgment Prediction \cite{LJPAAAI2020,malik-etal-2021-ildc,chalkidis-etal-2019-neural,Aletras2016,chen-etal-2019-charge,LongT2019,Xu2020,Yang2019,kapoor-etal-2022-hldc}, Rhetorical Role Labeling \cite{bhattacharya2019identification,saravanan2008automatic,malik-etal-2022-semantic,kalamkar-etal-2022-corpus}, Summarization \cite{Tran2019}, Prior Case Retrieval and Statute Retrieval \cite{COLIEE2021,rabelo2020coliee,kim2019statute}. Due to space limitations, we do not go into details of various tasks and allude to surveys on Legal-AI by \citet{westermann2022data,villata2022thirty,colonna2021reflections}. There have been some works in the area of Rhetorical Role Labeling and Legal Judgement Prediction; however, the research problems remain far from being solved, hence we organize shared tasks on these tasks. Moreover, Legal-NER (Named Entity Recognition) is an unexplored area, and we would like to draw the community's attention to it. 

%% file: tasks.tex
\begin{figure}[t]
\begin{center}
\includegraphics[scale=0.17]{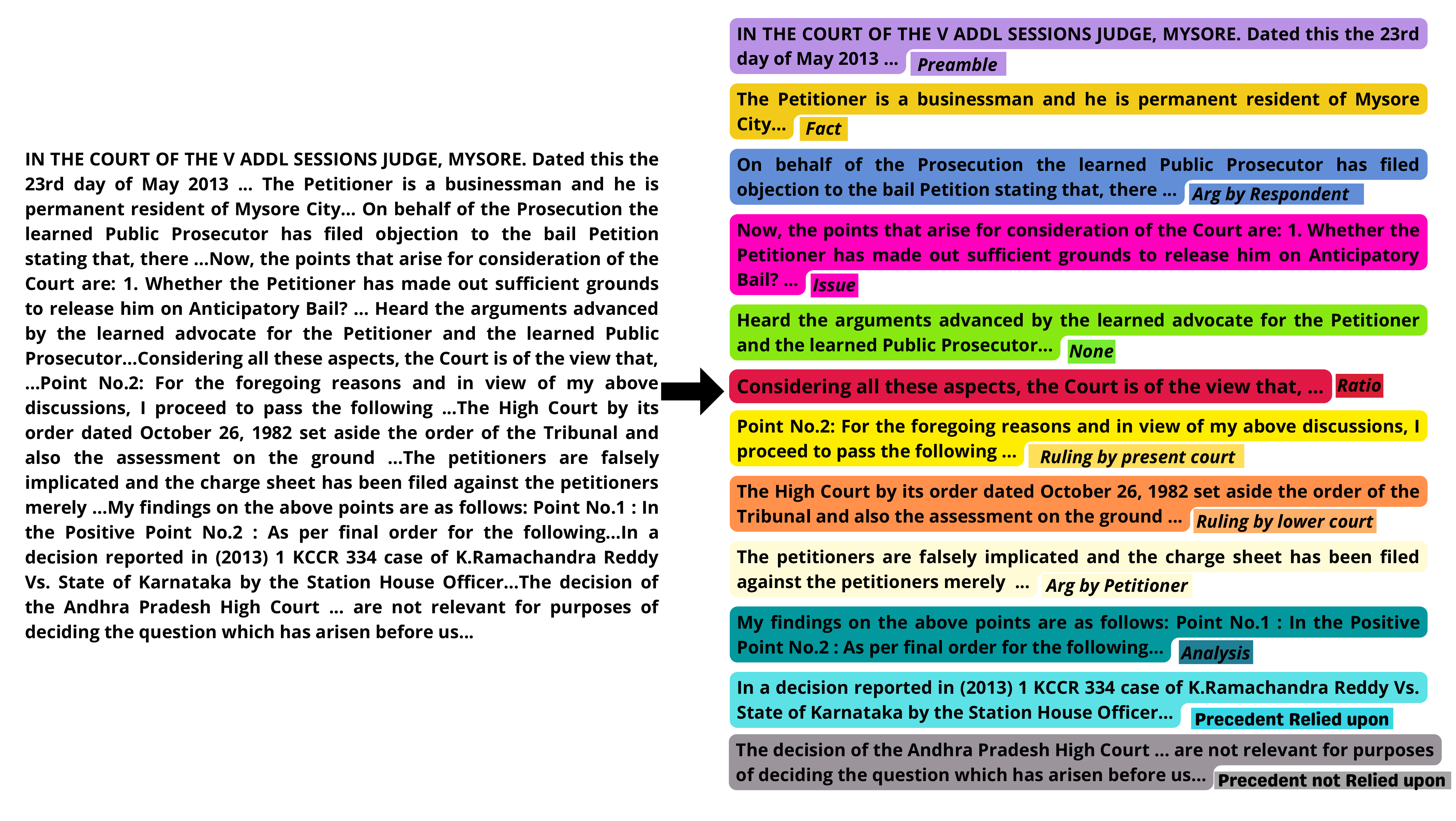}
\caption{Example of document segmentation via Rhetorical Roles labels. On the left is an excerpt from a legal document, and on the right is a document segmented and labeled with rhetorical role labels.}
\label{fig:rr}
\vspace{-5mm}
\end{center}
\end{figure}

%malik-etal-2022-semantic,kalamkar-etal-2022-corpus,kalamkar-etal-2022-named, malik-etal-2021-ildc

\section{Tasks Description} \label{sec:tasks}
In LegalEval 2023, we proposed the following three shared tasks, which focus on automatically understanding legal documents. The documents were in English. 

\noindent\textbf{Task A: Rhetorical Roles (RR) Prediction:} Given that legal documents are long and unstructured, we proposed a task for automatically segmenting legal judgment documents into semantically coherent text segments, and each such segment is assigned a label such as a preamble, fact, ratio, arguments, etc. These are referred to as Rhetorical Roles (RR). Concretely, we propose the task of Rhetorical Role Prediction: the task is to segment a given legal document by predicting (Figure \ref{fig:rr}) the rhetorical role label for each sentence \cite{malik2021RR,kalamkar-etal-2022-corpus}. We target 13 RRs as outlined in our previous work \citet{kalamkar-etal-2022-corpus}: Preamble, Facts, Ruling by Lower Court, Issues, Argument by Petitioner, Argument by Respondent, Analysis, Statute, Precedent Relied, Precedent Not Relied, Ratio of the decision, Ruling by Present Court and None. For the definition of the RRs, please refer to \citet{kalamkar-etal-2022-corpus}. 
%Due to space constraints, we do not describe the  RR label definitions in this proposal. %Figure \ref{fig:rr} shows an example of the RR prediction task.    

%An unstructured legal judgment document is segmented into semantically coherent parts, and each part is annotated with a rhetorical role label such as preamble, fact, ratio, etc. Each sentence in a legal document is annotated with a rhetorical role label in the proposed corpus. Typically, consecutive sentences can have a similar role in a judgment document. Task is to train AI model to predict the sentence rhetorical role. This task can be thought as sequential sentence classification in which rhetorical role of each sentence is dependent not only on the words of the current sentence but also the rhetorical roles of neighboring sentences. 

\noindent\textit{RR Task Relevance:} The purpose of creating a rhetorical role corpus is to enable an automated understanding of legal documents by segmenting them into topically coherent units (Figure \ref{fig:rr}). This segmentation is a fundamental building block for many legal AI applications like judgment summarization, judgment outcome prediction, precedent search, etc. %This task will provide the required data to build this fundamental building block.

\begin{figure}[t]
\begin{center}
\includegraphics[scale=0.9]{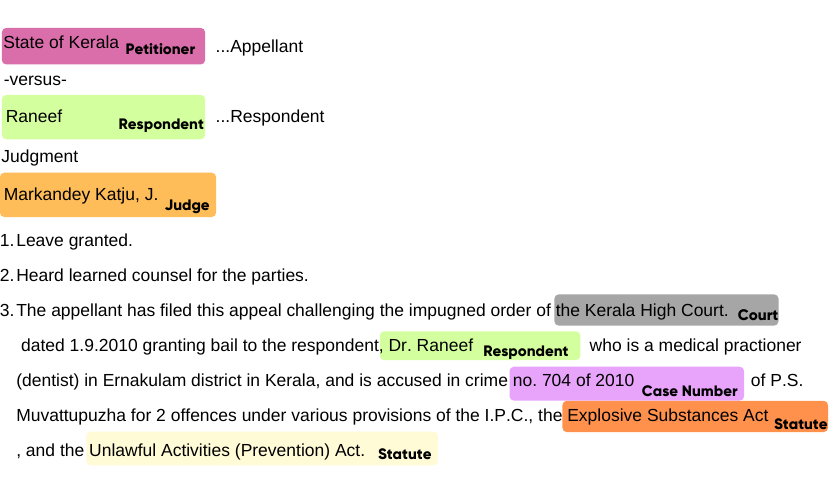}
\caption{Example of Named Entities in a court judgment}
\label{fig:ner}
\vspace{-7mm}
\end{center}
\end{figure}

\noindent\textbf{Task B: Legal Named Entity Recognition (L-NER):} Named Entity Recognition is a widely studied problem in Natural Language Processing \cite{https://doi.org/10.48550/arxiv.1812.09449,dozier2010named}. However, legal documents have peculiar entities like names of the petitioner, respondent, court, statute, provision, precedents,  etc. These entity types are not recognized by the standard Named Entity Recognizer. Hence there is a need to develop a Legal NER system; consequently, we proposed Task B as Legal Named Entity Recognition (L-NER). In this task, we target 14 legal named entities as described in our previous work \citet{kalamkar-etal-2022-named} and also shown in Table \ref{table:NER}. An example of the task is shown in Figure \ref{fig:ner}. 

\begin{table}
\tiny
\centering
\renewcommand{\arraystretch}{0.7}
\setlength\tabcolsep{0.5pt}
\begin{center}
\begin{tabularx}{\columnwidth}{lX}
    \toprule
        \textbf{Entity}  & \textbf{Description} \\ \midrule
        Court  & Name of the courts (supreme, high court, district, etc.) \\ \midrule
        Date  & Dates mentioned in the. Judgement \\ \midrule
        Police Station  & Name of Police Stations \\ \midrule
        Organization  & Name of organization apart from court \& police stations.  E.g., private companies, banks, etc. \\ \midrule
        Statute  & Act or law under which case is filed \\ \midrule
        Provision  & Sections, articles or rules under the statute \\ \midrule
        Precedent  & Higher Court cases referred \\ \midrule %They typically are supreme court cases and judgment refer to them \\ \hline
        Case number  & Number of the case \\ \midrule
        Petitioner Name & Names of Appellant/ Petitioner. \\ \midrule
        Respondent Name  & Names of Respondent \\ \midrule
        Judge Name  & Name of judge \\ \midrule
        Lawyer Name  & Name of Lawyers \\ \midrule
        Witness Name  & Name of witnesses \\ \midrule
        Others & Names of other people that don't belong to any other categories above \\ \bottomrule
    \end{tabularx}
    \caption{Legal Entity Types}
\label{table:NER}
     \end{center}
 \vspace{-5mm}
\end{table}

\noindent\textit{Task Relevance:}
Similar to the importance of NER in the general NLP domain, Legal NER is equally essential in the legal domain. For example, L-NER is the first step in extracting relevant entities for information extraction and retrieval-based tasks. Blackstone (\url{https://github.com/ICLRandD/Blackstone}) provides a library for L-NER; however, it is limited to only UK cases and does not generalize well to case documents from other countries; moreover, the set of entities covered is quite limited. 

\noindent\textbf{Task C: Court Judgment Prediction with Explanation (CJPE):} In recent years, in order to augment a judge in predicting the outcome of a case, the task of legal judgment prediction has been an active area of research \cite{malik-etal-2021-ildc}. However, given the nature of the legal process, judgment prediction without providing an explanation for the judgments is not of much practical relevance to legal practitioners. We proposed the task of Legal Judgement Prediction with Explanation, where, given a legal judgment document, the task involves automatically predicting the outcome (binary: accepted or denied) of the case and also providing an explanation for the prediction. The explanations are in the form of relevant sentences in the document that contribute to the decision. This task is an extension of one of our previous works \cite{malik-etal-2021-ildc}. \textbf{We  divided the task into two sub-tasks: (1) Court Judgment Prediction and (2) Explanations for the prediction}. As done in our previous work, we take all the steps to remove any bias in the dataset and address the ethical concerns associated with the task.  

\noindent\textit{Task Relevance:} The volume of legal cases has been growing exponentially in many countries; if there is a system to augment judges in the process, it would be of immense help and expedite the process. Moreover, if the system provides an explanation for the decision, it would help a judge make an informed decision about the final outcome of the case. 

\subsection{Evaluation}
The rhetorical roles task (a multiclass prediction problem) is evaluated using a weighted F1 score based on the test data.
For the CJPE task, the evaluation for judgment prediction (binary classification) is done using the standard F1 score metric, and for the explanation sub-task, we use ROUGE-2 score \cite{malik-etal-2021-ildc} for evaluating the machine explanations with respect to the gold annotations. For L-NER, we use standard F1 score metrics \cite{segura-bedmar-etal-2013-semeval}. Automatic evaluations of the submitted models are done using CodaLab, and a leaderboard is  maintained (\url{https://codalab.lisn.upsaclay.fr/competitions/9558}). 

%% file: corpus_annotations.tex
\section{Corpora} \label{sec:corpus}

We focus on freely and publicly available Indian legal documents in English. The documents are obtained from Indian Legal search portal IndianKanoon (\url{https://indiankanoon.org/}). India follows a common-law system, which is similar to the one in the U.S. and the U.K. The tasks organizers with the help of legal experts in the past, have created the annotated datasets for RR \cite{kalamkar-etal-2022-corpus,malik2021RR}, L-NER \cite{kalamkar-etal-2022-named} and CJPE \cite{malik-etal-2021-ildc}. The data is annotated by senior law students, professors, and legal practitioners, thus ensuring the high quality of the data. %For LegalEval, corpora were further expanded. %For the L-NER task, currently, we are in the process of annotating the documents with the help of law students and law professors. 
We annotated around $250$ judgments which are likely to have $850$k tokens and roughly $10$k entities. The RR data has about $265$ legal documents with $841,648$ tokens and $26,304$ sentences, with an average of $3176$ tokens per document. It is split into the train, validation, and test sets with an approximate 70-10-20 split. L-NER corpus has about $14,444$ judgment documents annotated with $17,485$ named entities belonging to $14$ categories. The CJPE corpus has about $34K$ documents, and these are split into the train, validation, and test sets. 

%% file: teams.tex
\section{Participating Systems} \label{sec:teams}

\begin{figure}[t]
\begin{center}
\includegraphics[width=\linewidth]{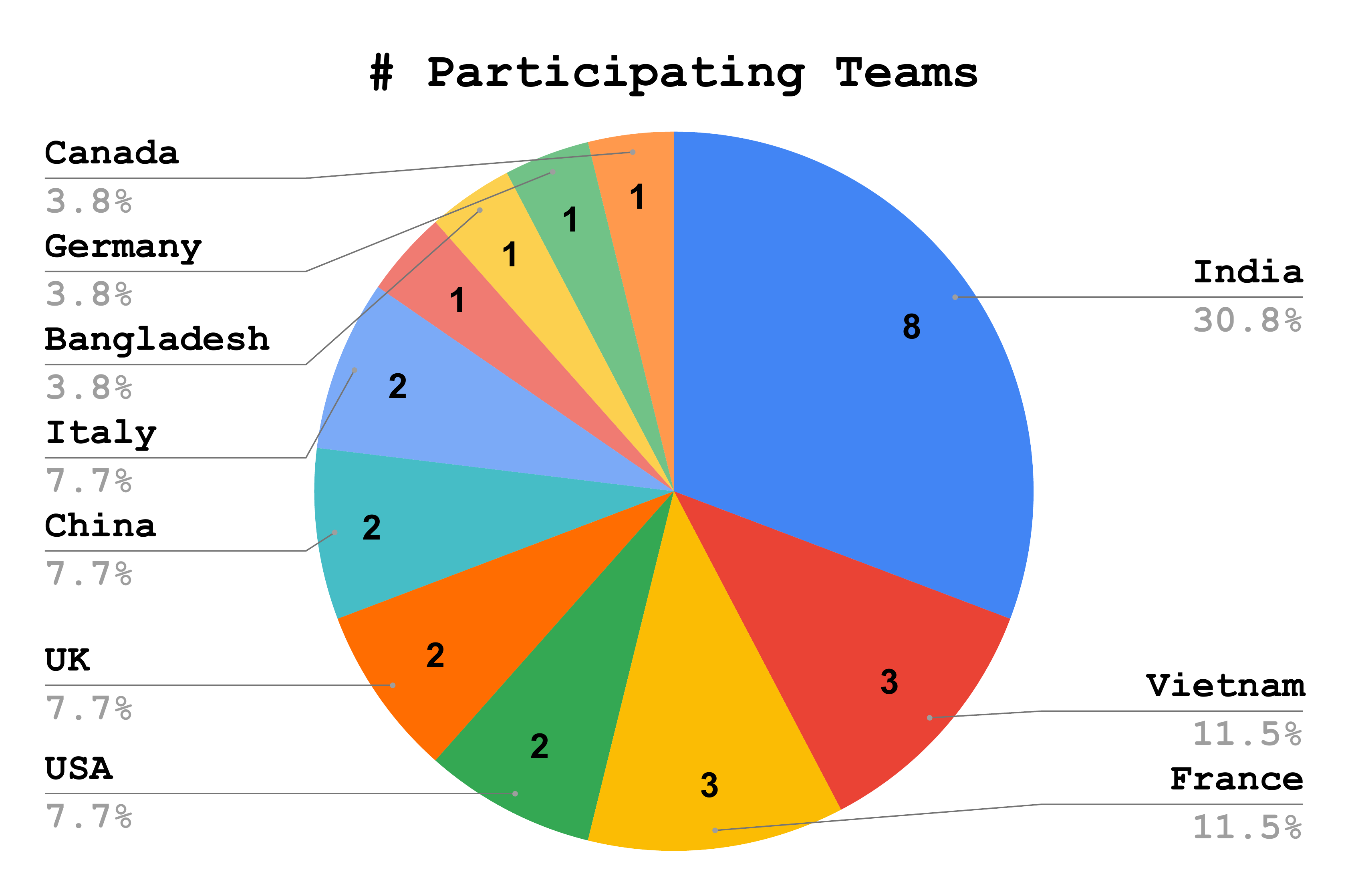}
\caption{Demographics of the participating systems}
\label{fig:demographics}
\vspace{-5mm}
\end{center}
\end{figure}

%\subsection{Participating Teams}
We organized the shared task using the CodaLab platform. In total, 40 teams participated in various sub-tasks and finally 26 teams (approx. 100 participants) submitted systems description papers. Overall, there were 17 submissions for Task-A (RR) and 11 submissions each for Task-B (NER) and Task-C (CJPE). Teams other than from India also participated in the shared task. The demographic distribution of teams is shown in Figure \ref{fig:demographics}. Next, we summarize the proposed approaches by various teams for each of the tasks. %The leaderboards for all three tasks are on the shared task website. 

\subsection{Task-A (Rhetorical Role Prediction)}
%\noindent \textbf{Task-A (RR): } 
In total, seventeen teams provided a description of their approach. We observed a wide variety of approaches for Task-A, like BiLSTM-based, Transformer-based, and GNN-based architectures, out of which BiLSTM was one of the popular choices (6 teams). Another common trend was the use of pre-trained transformer-based architectures as a backbone, e.g., BERT \cite{devlin-etal-2019-bert}, RoBERTa \cite{RoBERTa}, LegalBERT \cite{chalkidis-etal-2020-legal}, InLegalBERT \cite{paul2022pre}. 

%among which a popular choice was the use of BiLSTM layers in the proposed systems. Six out of seventeen teams proposed the incorporation of BiLSTM layers for capturing the contextual representation.

%We observed a wide variety of approaches for Task-1, among which a popular choice was the use of BiLSTM layers in the proposed architecture. In total, seventeen teams provided a description of the approach. Six out of seventeen teams proposed the incorporation of BiLSTM layers for capturing the contextual representation.

\texttt{AntContentTech} \cite{huo-EtAl:2023:SemEval} proposed a system based on Bi-LSTM with CRF layers and used domain-adaptive pre-training with auxiliary-task learning with pre-trained LegalBERT as a backbone. They also experimented with augmentation strategies like shuffling precedents among the documents and back translation. This helps to improve performance. 

%It is also interesting to notice that the various ways of augmentations, like shuffling precedents among the documents and back translation, turn out to be a good strategy and help improve the performance highlighting the significance of augmentations on legal texts to improve RR systems.

% ``AntContentTech" 
% "Bi-LSTM + CRF. They enhance the system with domain-adaptive pre-training, data augmentation, and auxiliary-task learning. LegalBERT(zlucia) is used as pre-trained model.
% One thing that they are balancing the dataset by shuffle blocks with the same class across different documents. For example, take precedent relied from another document and replace the one that is in this document. This is done in addition to back translation. Fundamentally it looks like that more data is used to train the model by augmenting it smartly.
% Incorporating Indian Legal text has performance gains"

\texttt{VTCC-NLP} \cite{nguyen-ngo-bui:2023:SemEval} followed a unique strategy of using Graph Neural Nets with sentence embeddings. Moreover, the proposed method used multitask learning with label shift prediction and contrastive learning to separate close labels like \textit{ARG\_petitioner} and \textit{ARG\_respondent}. Combining entity-based GNN (GAT \cite{velickovic2018graph}) with multitask learning and contrastive learning helped to improve performance.

% "RR Leaderboard 4th rank. Used Graph Neural Nets on top of sentence embeddings for contextual sentence embeddings + Multitask learning using Label shift prediction. Used contrastive learning to differentiate close labels like ARG\_petitioner, ARG\_respondent.
% Combination of mutitask learning + InLegalBERT + contrastive learning + entity-based GNN is improving the performance."

\texttt{TeamShakespeare} \cite{jin-wang:2023:SemEval} highlighted the significance of considering inter and intra-sentence representations in RR predictions. The method also reported improvements achieved by replacing BERTbase with LegalBERT as a backbone in the architecture.

% TeamShakespeare
% "Ranked 5th. Legal- BERT-HSLN model that considers the context information in both intra- and inter-sentence levels to predict rhetorical roles.
% Replacing the BERTbase with Legal BERT has improved the score"

\texttt{NLP-Titan} \cite{kataria-gupta:2023:SemEval} followed a two-stage Hierarchical strategy where in the first stage, similar classes were merged, and a sequence sentence classification algorithm based on Hierarchical Sequential Labeling Network (HSLN) was used. In the second stage, the SetFit model is trained over the fine-grained merged classes. 

% NLP-Titan	
% The authors propose a two-stage approach for classifying legal documents based on their Rhetorical Roles. In the first stage, similar classes are merged together and trained using a sequence sentence classification algorithm based on Hierarchical Sequential Labeling Network (HSLN). In the second stage, merged classes are trained separately with the assistance of the SetFit model. The authors use BERT-based word embeddings and a Bi-LSTM network to capture contextual meaning within the sentence and extract contextual information from surrounding sentences. A CRF layer is used for prediction. The method enables accurate and efficient classification of legal documents based on their specific Rhetorical Roles.

\texttt{YNU-HPCC} \cite{chen-EtAl:2023:SemEval3} proposed the use of LegalBERT with attention pooling and a CRF module. Predictions are made using the Deep ensembles \cite{deep_ensembles} strategy, where multiple models were trained with different seeds followed by a voting scheme.

% YNU-HPCC	Legal BERT + attention pooling + CRF. Multiple models with same architecture and different seed   were trained and voting was used to come up with final inference.

\texttt{TeamUnibo} \cite{noviello-EtAl:2023:SemEval} used a pretrained BERT with two BiLSTM layers followed by average pooling and a linear layer. The proposed method was motivated to maintain topical coherence by taking the surrounding sentences as context for RR classification.

% TeamUnibo
% The approach for the Rhetorical Role segmentation task followed a typical NLP pipeline. A context window was created for each sentence during preprocessing. BERT and RoBERTa were used as baseline models for sentence classification, but a model was proposed that utilized a pre-trained BERT embedding module followed by two Bi-LSTM layers and a linear classifier, with average pooling at the token level. Two models, Legal-RoBERTa and InLegalBert, were evaluated using this architecture and fed with the target sentence and its surrounding context to generate contextually enriched sentence embeddings, which were then fed into a linear classifier to output class predictions. The model takes advantage of the structure of legal judgments, where sentences tend to maintain a topical coherence, to reason on the context of the sentences when classifying.

\texttt{IRIT\_IRIS\_(A)} \cite{lima-moreno-aranha:2023:SemEval} proposed a unique style of segmenting long documents referred to as DFCSC (Dynamic Filled Contextualized Sentence Chunks), where every chunk contained a variable number of core and edge sentences. The paper further described 3 variants of the proposed method and reported the results.

% "This paper proposes Dynamic Filled Contextualized Sentence Chunks (DFCSC).These are chunks with a variable number of core and edge sentences, without truncated core sentences, and that usually do not require padding tokens
% They employed two sets of models relying on DFCSCs:
% 1. DFCSC-SEP: Three models rely on DFCSC for sentence embeddings. InCaseLaw, Roberta-base and Longformer. In all these models, core sentences are represented by the hidden states of the respective separator tokens.
% 2.DFCSC-CLS :each core sentence is represented by the concatenation of hidden states of the CLS and SEP tokens.
% The best model achieved an F1 score of 0.8139.
% New style of segmenting long documents into contextualized chunks - DFCSC"

\texttt{LRL\_NC} \cite{tandon-chatterjee:2023:SemEval2} used a hierarchical sequence encoder with contextual sentence representations followed by a CRF layer for classifying sentences into RR categories.

% LRL\_NC	
% "They use a hierarchical sequence encoder consisting of basic sentence representation and a contextualised sentence representation followed by a CRF layer for final label prediction
% HDP features improve the overall performance of the system."

\texttt{ResearchTeam\_HCN} \cite{singh-EtAl:2023:SemEval1} used back translations to balance the imbalanced classes and proposed using attention pooling with sentence transformers.  

% ResearchTeam\_HCN	
% "They use translation back and forth in another language similar to english to balance the imbalanced class.They use attention pooling for their Sentence Transformer model.
% They say that choosing 512 instead of 256 embedding size gives better results. They repeat that due to class imbalance prediction of classes which appear lower compared to others is where model is failing to classify and hypothesise that enriching embedding and adding a CRF layer for classifier would improve the results."

\texttt{UOttawa.NLP23} \cite{almuslim-EtAl:2023:SemEval} highlighted improvements with a Hierarchical BiLSTM-CRF when compared to other transformer-based models.

% uottawa.nlp23	
% Authors have performed experiments with various models architectures and finally uses a Hierarchical BiLSTM-CRF model, which in their experiments outperforms the transformer-based models. The experiment uses four different embeddings and finds that randomly initialised word embeddings work best with the Hierarchical BiLSTM-CRF model. The transformer-based models perform poorly on the test dataset, while performing well on the validation dataset.

\texttt{NITK\_Legal} \cite{sindhu-EtAl:2023:SemEval} also used a BiLSTM architecture with the CRF layer, which follows a hierarchical structure. The paper also reported its findings on various sentence encoding approaches like BiLSTM embeddings, sentence2vec (sent2vec) \cite{moghadasi2020sent2vec}, Universal Sentence Encoder \cite{cer2018universal}, Sentence BERT (SBERT) \cite{reimers2019sentence}, trigram embeddings,  weighted trigram and handcrafted features + sent2vec and reported the importance of handcrafted features along with sent2vec embeddings for RR classification.
 
% nitk\_legal	
% "In the proposed approach authors have used Hierarchical BiLSTM architecture with the CRF layer deployed on top of it to automatically extract the sequence of rhetorical roles for the provided judgement case. In their approach authors have used various approached to generate embeddings that is fed to BiLSTM for classification. For embedding generation approached used are: BiLSTM embeddings, sentence2vec, Universal Sentence Encoder, sentence BERT, Trigram Embeddings, Mean Trigram, Weighted Trigram and Handcrafted Features + sent2vec.
% Among all the approached used Handcrafted Features + sent2vec gave the best results in classification of rhetorical roles"

\texttt{T\"ueReuth-Legal} \cite{manegold-girrbach:2023:SemEval} explored the learning of a continuous distribution over positions in the document to indicate the likelihood of respective RRs appearing at certain positions. The method further incorporated domain knowledge through hard-coded lexical rules to improve performance.

% TueReuth-Legal
% The method described combines information from individual sentences and entire documents to improve performance of models that have limited context. This is achieved by fine-tuning a pre-trained language model (Roberta-base, Legal-BERT to predict RRs from individual sentences, learning a conditional distribution of RRs given all RRs predicted so far, learning a binary score for all ordered pairs of RRs to indicate allowed transitions from the previous RR to the following, learning a continuous distribution over positions in the document to indicate the likelihood of respective RRs appearing at certain positions, and applying domain knowledge through hard-coded lexical rules.

\texttt{NITS\_Legal} \cite{jain-borah-biswas:2023:SemEval} created a different version of the dataset for capturing local context and used a BiLSTM-based sentence classification approach, highlighting the significance of context in RR prediction.

% NITS\_Legal	
% "Author proposed a BiLSTM based sentence sequence labeling approach that uses a local context-incorporated dataset created from the original dataset. To better represent the sentences during training, authors have extracted legal domain specific sentence embeddings from Legal BERT model. Authors experimental findings emphasizes the importance of considering local context instead of treating each sentence independently to achieve better performance in this task.
% "

\texttt{Steno AI} \cite{gupta-EtAl:2023:SemEval} created a graphical structure for capturing the relationship between the sentences in the document and applied GNN-based classification to predict the rhetorical roles. 

% Steno AI	
% The paper discusses about two approaches for the RR task, Graph-based approach and Context-based Legal-BERT. In the graph based approach, the CLS tokens for each sentence are extracted from the trained Legal-BERT and cosine similarity is calculated between the pairs. If the value is greater than 0.5 then an edge is created with the cosine similarity acting as edge weight. Label diffusion algorithm is performed on this graph to classify sentences into similar category. In another graph based approach, Legal-Bert embeddings for each sentence are used as node representations and a two-layer Graph Convolution Network (GCN) is used to perform classifications on the data. In the context-base Legal BERT approach, each sentence is appended with two preceding sentences and two succeeding sentences. These 5 sentences are fed as input to Legal-BERT and default classifier performed classification on these context-based inputs.

\texttt{CSECU-DSG} \cite{tasneem-EtAl:2023:SemEval}	framed the task as a multi-class sequence classification task. In the proposed approach the authors utilized the fine-tuned Legal-BERT for classifying legal sentences into 13 rhetorical roles.

\texttt{LTRC} \cite{baswani-sriadibhatla-shrivastava:2023:SemEval} used several algorithms to classify sentence embeddings into different classes of Rhetorical Roles. The authors showed that using 512 instead of 256 as embedding size gives better results. Further, they showed that enriching the embeddings and adding a CRF layer to the classifier helps to improve results for classes with fewer samples.

%"They have used several algorithms to classify sentence embeddings into different classes of Rhetorical Roles. They say that choosing 512 instead of 256 embedding size gives better results. They repeat that due to class imbalance prediction of classes which appear lower compared to others is where model is failing to classify and hypothesise that enriching embedding and adding a CRF layer for classifier would improve the results."

\texttt{UO-LouTAL} \cite{bosch-EtAl:2023:SemEval} proposed a Bag-of-Words-based system and used NER labels to replace entities with their types to reduce sentence variability.

% UO-LouTAL	
% "The approach aims to identify Rhetorical Roles (RR) using a Bag-of-Words system and a Deep Learning vector-based approach. The input sentences were first labeled with an NER system, and the entities were replaced by their type to reduce sentence variability and enrich them with semantic information. The Bag-of-Words system was developed using the scikit-learn library, with a combination of TF-IDF weights and tested out uni- and bi-grams. The Deep Learning vector-based approach used the SBERT sentence transformer to convert each datapoint into a sentence vector and fed into a K-Neighbors model. The approach also includes a rule-based post-processing. In rule based two main rule are use one is for identifying end of preamble with the help of keyword and a 3 sentence sliding window is used to identify association of sentences to same RR category.
% "

\subsection{Task-B (Named Entity Recognition)}
%\noindent\textbf{Task-B (NER):}
For the NER task, out of 17 participating teams, 11 teams submitted system description papers. Most teams (7) used variations of the BERT-based transformers like RoBERTa, DeBertaV3\cite{debertav3}, LegalBERT, Albert \cite{albert}. techniques based on XLNet \cite{xlnet}, LUKE-base, and Legal-LUKE are also implemented. BiLSTM and CRF layers are used in the final layers on the top of transformers. 

\texttt{ResearchTeam\_HCN} \cite{singh-EtAl:2023:SemEval1} proposed two baseline models and one main model with an MTL (Multi-Task Learning)  framework. They used data augmentation (back-translation) as a way to make up for the imbalance in class data. Further, they used a general-purpose NER system to aid the Legal-NER model. For example, using rule-based techniques they assigned labels to an entity predicted by general-purpose NER. %The technique is seemingly standard i.e. translate from English to a language with similar grammar and vocabulary like German or Dutch and then back translated to get generate another example of a class then it is put back in the original text with other classes to create a balanced dataset.   
%A generic NER is used to fix the Legal NER. For example, sometimes a person isn't tagged properly as Witness. But in generic NER it gets detected as person. So based on context and some similarity score, we can assign it to some class. They use some rule based techniques to further augment the response.

\texttt{VTCC-NER} \cite{tran-doan:2023:SemEval} proposed the use of ensembles to improve the NER predictions via two approaches. The first approach considered the use of multiple transformer backbones (RoBERTa, BERT, Legal-BERT, InLegalBERT, and XLM-RoBERTa) in the spaCy framework to make the predictions, which are further combined using a weighted voting scheme to generate the NER predictions. In the second approach, they  adapted MultiCoNER (Multilingual Complex Named Entity Recognition \cite{malmasi2022semeval}) with multiple pre-trained backbones to generate the contextualized features, which are further concatenated with dependency parsing features and part-of-speech features. The obtained concatenated features were further passed through a CRF module to generate the predictions. %The first approach with weighted ensembles performs better, with a score of 90.9873 on the test set, ranking 2nd on the NER leaderboard.

\texttt{Jus Mundi} \cite{cabreradiego-gheewala:2023:SemEval} proposed the method FEDA (Frustratingly Easy Domain Adaptation), and trained it over multiple legal corpora. The FEDA layer consisted of a GeLU activation layer, followed by a linear layer and layer normalization. The method extracted features from DeBERTa V3 and passed them through two splits: the general FEDA layer and the specialized FEDA layer. The specialized FEDA layer also incorporated features from the general FEDA layer, and the final output was passed through a CRF module to generate the predictions. %The paper further provides analysis and findings of various submissions made for the task highlighting the role of the voting scheme and post-processing with minor improvements from 0.8908 to 0.9007.

\texttt{UOttawa.NLP23} \cite{almuslim-EtAl:2023:SemEval} converted the text to B-I-O format and tokenized it using the Legal-Bert \cite{chalkidis-etal-2020-legal}. The authors proposed a token classification approach where each token was classified into a particular entity type in the dataset.

\texttt{TeamUnibo} \cite{noviello-EtAl:2023:SemEval} presented an approach to handle raw text without any pre-processing. A sliding window approach handled long texts and avoided input truncation. The system used the B-I-O labeling format and mapped labels from words to tokens. Several baseline models, including RoBERTa, InLegalBERT, and XLNet, were tested, along with two custom models that use RoBERTa or XLNet as the transformer backbone, followed by a BiLSTM and CRF layers. The best-performing model was found to be RoBERTa - BiLSTM - CRF and XLNet - BiLSTM - CRF.

\texttt{AntContentTech} \cite{huo-EtAl:2023:SemEval} used a SciBERT-HSLN to get word embeddings of each sentence which were then passed through a Bi-LSTM layer and an attention-based pooling layer to get sentence representations. These representations are directly fed into a CRF layer to perform the span classification. %Compared to the baseline performance, the gain is less than 1.5\%.

\texttt{PoliToHFI} \cite{benedetto-EtAl:2023:SemEval}	applied the entity-aware attention mechanism implemented in the LUKE model \cite{yamada-etal-2020-luke}. %While this method has not even beaten the benchmark, The BERT models pre-trained on legal data showed improvement in performance over their counterparts.

\texttt{Ginn-Khamov} \cite{ginn-khamov:2023:SemEval} proposed a three step augmented approach. First, a classification model, where the sentence class was added as an input token, then a twin transformer model with two identical NER models was used and each batch of sentences was split into preamble and judgment using the previous classification model. The sentence class was used to pick one of the two sub-models, and finally, a modified token classification linear layer took the sentence class provided by the classifier predictions to learn the differences between documents.

\texttt{TeamShakespeare} \cite{jin-wang:2023:SemEval} introduced Legal-Luke, a legal entity recognition model based on the bidirectional transformer encoder of LUKE \cite{yamada-etal-2020-luke}. The model generated the Legal contextual embeddings for words and entities by adding the token embeddings, type embeddings. Position embeddings were added to compute the word and entity embeddings.

\texttt{UO-LouTAL} \cite{bosch-EtAl:2023:SemEval} aimed to achieve  interpretability in the L-NER task by using a standard CRF. The input to the CRF model was a structure containing token features retrieved from the text using SpaCy, and the features of its preceding and succeeding tokens. Optionally, post-processing was applied to extract specific patterns such as dates or case numbers.

\texttt{Legal\_try} \cite{zhao-EtAl:2023:SemEval1} proposed the use of multiple heterogeneous models, which were further selected using a recall score. Later, all the selected models were used in a voting scheme to generate the predictions. The heterogeneous models included ALBERT – BiLSTM – CRF, ALBERT – IDCNN – CRF, and RoBERTa (dynamic fusion) – BiLSTM/IDCNN – CRF.

\texttt{Nonet} \cite{nigam-EtAl:2023:SemEval} used a combination of results produced by the custom-trained spaCy model and fine-tuned BERT-CRF model for the final prediction. spaCy was trained by replacing the RoBERTa embeddings with the law2vec embeddings and the LegalBERT was used in the pipeline for fine-tuning the model. The BERT-CRF model was trained over the pre-processed text using the BIO format. If the span of the entities predicted by both models intersected then the entity with a larger span was considered. For the rest, a union of the result produced by both models was considered.
%\AM{Kiran to complete team Nonet}

\subsection{Task C1 and C2: Legal Judgment Prediction and Legal Judgement Prediction with Explanation}
%\noindent\textbf{Task C1 and C2: LJP and CJPE}
%\AM{Kiran to write a summary for both}
For the Legal Judgement Prediction task (C1) transformers were used as the base for 7 out of 8 participating systems. Longformer, RoBERTa, InLegalBERT, LegalBERT, and CaseLawBERT are used. One team \texttt{nclu\_team} proposed the use of CNN to solve this task. In the Explanation for Judgement Prediction task (C2), various techniques were employed to generate the explanations. Some of those included the use of occlusion, clustering of similar sentences and picking sentences from the clusters, using relevance scores based on the proportion of legal entities,  use of attention scores, extractive summarization, span lengths, and generation using the T5-based transformer model. 
%\texttt{Viettel-AI} \cite{Viettel-AI} 
%highlights the effectiveness of using LongFormer to process larger token sizes and considers the first or last chunk to represent the document. 

\texttt{Viettel-AI} \cite{hoang-bui-bui:2023:SemEval} used LongFormer \cite{beltagy2020longformer} to process documents. High and low-frequency words were filtered out and a TF-IDF representation of the document was created. The TF-IDF vector representation was used along with LongFormer to learn representations in an end-to-end fashion. The authors further extended the proposed method for explanations by using a trained judgment prediction model and masking the sentences present in the document. With the help of this occlusion technique, the authors ranked the sentences based on the difference in the judgment prediction probability scores.

\texttt{IRIT\_IRIS\_(C)} \cite{prasad-boughanem-dkaki:2023:SemEval} proposed a multi-level encoder-based classifier using a pretrained BERT-based encoder as a backbone and fine-tuned over the chunked version of the documents. Further, the extracted chunk embeddings were used to represent the entire document, followed by another transformer and recurrent layer in level-3 to learn the relation between the chunk representations. For explanations, the proposed method followed the idea of occlusion sensitivity as an extension to the judgment prediction module.
% . \AM{how is explanation done?}

% \AM{confusion between names and both look same}
% \AM{need to be completed by Abhinav}

% \texttt{IRIT\_IRIS\_(A)} \cite{lima-moreno-aranha:2023:SemEval}	"Ranked 2nd (macro-F1 metric) for its sub-task C-1 and 7th (ROUGE-2 metric) for sub-task C-2.
% Three-level encoder-based classification architecture that works by fine-tuning a BERT-based pre-trained encoder and post-processing the embeddings extracted from its last layers, using transformer encoder layers and RNNs.
% Lack of case documents from these recent dates in test data, the models trained on the train set will not be able to predict properly due to a lack of learning from the new law articles or case proceedings."

\begin{table*}[t]
%\tiny
\small
\centering
\renewcommand{\arraystretch}{1}
\setlength\tabcolsep{1pt}
\begin{tabular}{ccccc}
\toprule
Rank & Team Name         & Paper Submission & Model                                                                 & Score  \\
\midrule
1    & \texttt{AntContentTech}    & yes                   & LegalBERT+BiLSTM+CRF                                                 & 0.8593 \\
2    & x\texttt{ixilu556}         & no                    &                                                                       & 0.8581 \\
3    & \texttt{HJ}                & no                    &                                                                       & 0.8497 \\
4    & \texttt{VTCC-NLP}          & yes                   & InLegalBERT + contrastive learning + entity-based GNN                 & 0.8389 \\
5    & \texttt{TeamShakespeare}   & yes                   & Legal + BERT + HSLN                                                      & 0.8343 \\
6    & \texttt{NLP-Titan}         & yes                   & HSLN + BERT + CRF                                                      & 0.8309 \\
7    & \texttt{YNU-HPCC}          & yes                   & Legal-BERT + attention pool + CRF                                  & 0.8146 \\
8    & \texttt{TeamUnibo}         & yes                   & InLegalBERT + BiLSTM                                                 & 0.8112 \\
9    & \texttt{IRIT\_IRIS\_(A)} & yes                   & DFCSC ( Dynamic Filled Contextualized Sentence Chunks)                & 0.8076 \\
10   & \texttt{peanut}            & no                    &                                                                       & 0.8038 \\
11   & \texttt{DeepAI}            & no                    &                                                                       & 0.8028 \\
12   & \texttt{LRL\_NC}           & yes                   & RoBERTa + HDP + BiLSTM + CRF                                          & 0.7980  \\
13   & \texttt{NU}                & no                    &                                                                       & 0.7951 \\
14   & \texttt{ResearchTeam\_HCN} & yes                   & Sentence Transformers + BiLSTM                                        & 0.765  \\
15   & \texttt{UOttawa.NLP23}     & yes                   & Hierarchical BiLSTM-CRF                  & 0.7465 \\
16   & \texttt{nitk\_legal}       & yes                   & Handcrafted Features + sent2vec          & 0.7400   \\
17   & \texttt{TueReuth-Legal}    & yes                   & RoBERTa-Base / Legal- BERT + temporal distribution of labels & 0.7335 \\
18   & \texttt{NITS\_Legal}       & yes                   & LegalBERT+BiLSTM                                                      & 0.7143 \\
19   & \texttt{Steno AI}          & yes                   & Context-based LegalBERT (best results)                                & 0.7102 \\
20   & \texttt{scholarly360}      & no                    &                                                                       & 0.6902 \\
21   & \texttt{aashsach}          & no                    &                                                                       & 0.6602 \\
22   & \texttt{CSECU-DSG}         & yes                   & LegalBERT                                                             & 0.6465 \\
23   & \texttt{LTRC}              & yes                   & LegalBERT+LDA / InCaseLawBERT                                                          & 0.6133 \\
% 23   & \texttt{LTRC}              & yes                   & LegalBERT+LDA(CLS/meanpool) / InCaseLawBERT(CLS/meanpool)                                                          & 0.6133 \\
24   & \texttt{ccidragons}        & no                    &                                                                       & 0.5813 \\
25   & \texttt{UO-LouTAL}         & yes                   & TF-IDF + SBERT                                                        & 0.5681 \\
26   & \texttt{achouhan}          & no                    &                                                                       & 0.3506 \\
27   & \texttt{viniciusvidal}     & no                    &                                                                       & 0.2609 \\
\bottomrule
\end{tabular}
\caption{Leaderboard Results for Task-A: Rhetorical Roles (RR) Prediction}
\label{table:Task-1-Results}
\vspace{-5mm}
\end{table*}

\texttt{UOttawa.NLP23} \cite{almuslim-EtAl:2023:SemEval} explored the performance of different transformer-based models and hierarchical BiLSTM-CRF models for legal text classification on the Indian Legal Document Corpus (ILDC). Different models, including Legal-BERT, CaseLawBERT, and InCaseLawBERT were fine-tuned and used as classifiers. The best-performing model was CaseLawBERT + BiGRU. The successful performance of CaseLawBERT over InCaseLawBERT can be attributed to the pretraining corpus used. However, the models struggled to generalize to longer texts, and the validation results were better than the test results due to the longer texts in the test set. The authors used a masking technique with CaseLawBERT and BiGRU models to extract explanations for legal outcomes. They clustered similar sentences and picked one or two sentences from each cluster to form explanations using two different approaches. The authors concluded that their low score was due to selecting only one/two sentences per chunk, which resulted  in the lower overlap.

%The best performing model was CaseLawBERT + BiGRU, achieving an F1-score of 0.68 on sub-task C-1, and the combination of transformers with sequential models showed potential in legal text classification tasks. The successful performance of CaseLawBERT over InCaseLawBERT can be attributed to the pretraining corpus used. However, the models struggled to generalize to longer texts, and the validation results were better than the test results due to the longer texts in the test set.

\texttt{PoliToHFI} \cite{benedetto-EtAl:2023:SemEval}	used a three-staged approach where in the first stage, 4 transformer based encoders were employed to find the section exerting the most significant influence on the classification performance, and the best-performing encoder was identified. In the second stage, document representation was obtained by average pooling the sentence representations obtained using the chosen model. This representation was passed to the linear layers to compute the final prediction. Finally, an ensemble prediction for the test set was generated from the combination of hierarchical models. They leveraged post hoc input attribution explanation methods to generate prediction explanations. The explanation was derived by adopting the post hoc attribution methods that  leverage the input document and the model. Further, a relevance score based on each sentence’s proportion of legal entities was assigned to boost the explanations. %post ad-hoc attribution method.

\texttt{ResearchTeam\_HCN} \cite{singh-EtAl:2023:SemEval1} followed a hierarchical model approach in which a Legal pre-trained RoBERTa model was fine-tuned using the last 512 tokens from each document. Later, the features were extracted using the fine-tuned Legal-RoBERTa model and were passed through the BiGRU model and a classification layer. %This approach has achieved the F1 score of 66.71\% for this task
For explanation, they generated relevant sentences using the T5-based transformer model and considered the task as a SEQ2SEQ approach. %On the given dataset, this method has achieved around 45 percent.

\texttt{AntContentTech} \cite{huo-EtAl:2023:SemEval} used a self-attention layer to generate contextualized sentence representations. A global token pre-pended to the input sequence was utilized to generate the final prediction of the judgment. The attention scores were used to extract the explanations. Based on these attention scores provided by the first token, the top $30$\% of sentences were selected.

%Other experiments using continued training shows that LegalBert$_{20w}$ \AM{what is this?} achieves an F1 score of 70.58 leading to a significant improvement in performance compared to F1 score of 68.16 achieved using LegalBert$_{zlucia}$. 
%uses a self-attention layer instead of the Bi-LSTM to generate contextualized sentence representations.The attention scores are used to extract the explanations. Based on these attention scores provided by the first token, top 30\% of sentences are selected. This method achieved a ROUGE-2 of 0.0445 in this task.

\begin{table*}[t]
%\tiny
\small
\centering
\renewcommand{\arraystretch}{1}
\setlength\tabcolsep{2pt}
\begin{tabular}{ccccc}
\toprule
Rank & Team Name         & Paper Submission & Model                                                  & Score  \\
\midrule
1    & \texttt{ResearchTeam\_HCN} & yes        & RoBERTa-base                                           & 0.9120  \\
2    & \texttt{VTCC-NER}          & yes        & RoBERTa + Transition based Parser with dynamic weights & 0.9099 \\
3    & \texttt{DeepAI}            & no         &                                                        & 0.9099 \\
4    & \texttt{Jus Mundi}         & yes        & DeBERTa V3 + FEDA + CRF                                & 0.9007 \\
5    & \texttt{Autohome AI}       & no         &                                                        & 0.8833 \\
6    & \texttt{uottawa.nlp23}     & yes        & LegalBERT                                              & 0.8794 \\
7    & \texttt{TeamUnibo}         & yes        & XLNet + BiLSTM + CRF                                   & 0.8743 \\
8    & \texttt{DamoAI}            & no         &                                                        & 0.8627 \\
9    & \texttt{AntContentTech}    & yes        & LegalBERT + CRF ensemble                               & 0.8622 \\
10   & \texttt{xixilu556}         & no         &                                                        & 0.8611 \\
11   & \texttt{PoliToHFI}         & yes        & LUKE-base / BERT-base                                  & 0.8321 \\
12   & \texttt{Ginn-Khamov}      & yes        & RoBERTa + Sentence class token                         & 0.7265 \\
13   & \texttt{TeamShakespeare}   & yes        & legal- LUKE                                            & 0.6670  \\
14   & \texttt{UO-LouTAL}         & yes        & SpaCy (preprocessing) + CRF                            & 0.6489 \\
15   & \texttt{Nonet}             & yes         &  custom trained spaCy + BERT-CRF                                                      & 0.5532 \\
16   & \texttt{Legal\_try}        & yes        & ALBERT / RoBERTa + BiLSTM / IDCNN +CRF                 & 0.5173 \\
17   & \texttt{shihanmax}         & no         &                                                        & 0.0186 \\
\bottomrule
\end{tabular}
\caption{Leaderboard Results for Task B: Legal Named Entity Recognition (L-NER)}
\label{table:Task-2-Results}
\vspace{-5mm}
\end{table*}

\texttt{nclu\_team} \cite{rusnachenko-markchom-liang:2023:SemEval} proposed solving judgment prediction via Convolution Neural Network and used attention layers over CNN to generate the explanations. The proposed use of a sliding window over the most attentive part is unique in terms of using the ranking order of the obtained attention scores and it provided better and comparable results when compared to other explanation approaches. 
% <Fill the summary here using the paper directly> \AM{need to be completed by Abhinav}

%Explanation as extractive summarization. treating document endings at first results in a 2.1\% improvement in judgment prediction across all the models. The inversion of the text documents and reduction of non-salient sentences (v2) using pairwise mutual information allows an improvement of the past result with extra $ \approx  3-4\%$"

\begin{table*}[t]
%\tiny
\small
\centering
\renewcommand{\arraystretch}{1}
\setlength\tabcolsep{1pt}
\begin{tabular}{ccccc}
\toprule
Rank & Team Name         & Paper Submission & Model                                                               & Score  \\
\midrule
1    & \texttt{Viettel-AI}        & yes        & LongFormer + TF-IDF                                                 & 0.7485 \\
2    & \texttt{IRIT\_IRIS\_(C)} & yes        & InLegalBERT+TransformerEncoder                                                 & 0.7228 \\
3    & \texttt{uottawa.nlp23}     & yes        & CaseLawBERT + BiGRU                                                 & 0.6782 \\
4    & \texttt{PoliToHFI}         & yes        & Transformer-based encoders + Avg. pool on segmented sentences & 0.6771 \\
5    & \texttt{maodou}            & no         &                                                                     & 0.6735 \\
6    & \texttt{ResearchTeam\_HCN} & yes        & RoBERTa + BiGRU                                                     & 0.6671 \\
7    & \texttt{AntContentTech}    & yes        & LegalBERT                                                         & 0.6416 \\
8    & \texttt{nclu\_team}        & yes        &   CNN + Attention + Sliding Window                                                                 & 0.6351 \\
9    & \texttt{DamoAI}            & no         &                                                                     & 0.6326 \\
10   & \texttt{xixilu556}         & no         &                                                                     & 0.6325 \\
11   & \texttt{Nonet}             & yes         &         BERT / Hierarchial Transformer + BiGRU+Attention                                                         & 0.5287 \\
\bottomrule
\end{tabular}
\caption{Leaderboard Results for Task-C-1: Court Judgment Prediction }
\label{table:Task-3a-Results}
\vspace{-3mm}
\end{table*}

\begin{table*}[t]
%\tiny
\small
\centering
\renewcommand{\arraystretch}{1}
\setlength\tabcolsep{5pt}
\begin{tabular}{ccccc}
\toprule
Rank & Team Name         & Paper Submission & Model                                                                                                    & Score  \\
\midrule
1    & \texttt{Nonet}             & yes         &       Keyword-based matching techniques + RR   & 0.5417 \\
2    & \texttt{Viettel-AI}        & yes        & Longformer + TF-IDF                                                                                      & 0.4797 \\
3    & \texttt{nclu\_team}       & yes        &    CNN + Attention + Sliding Window                                                                                                       & 0.4789 \\
4    & \texttt{xixilu556}         & no         &                                                                                                          & 0.4781 \\
5    & \texttt{DamoAI}            & no         &                                                                                                          & 0.4781 \\
6    & \texttt{uottawa.nlp23}     & yes        & CaseLawBERT + BiGRU                                                                                      & 0.4781 \\
7    & \texttt{PoliToHFI}         & yes        & post-hoc input attribution  & 0.4600   \\
8    & \texttt{ResearchTeam\_HCN} & yes        & T5 (modelled as a SEQ2SEQ approach)                                                                      & 0.4583 \\
9    & \texttt{AntContentTech}    & yes        & LegalBERT                                                                                             & 0.4525 \\
10   & \texttt{IRIT\_IRIS\_(C)} & yes        & InLegalBERT+TransformerEncoder                                                                                       & 0.4000    \\
11   & \texttt{maodou}            & no         &                                                                                                          & 0.3600  \\
\bottomrule
\end{tabular}
\caption{Leaderboard Results for Task-C-2:  Explanations for the Court Judgment Prediction}
\label{table:Task-3b-Results}
\vspace{-5mm}
\end{table*}

\texttt{Nonet} \cite{nigam-EtAl:2023:SemEval} used transformers like XLNet, Roberta, LegalBERT, InLawBert, InCaseLaw, and BERT-Large and also tried Hierarchical Transformer
model architecture with a moving window approach to reduce the document into chunks. The CLS representation of these chunks was then used as input to sequential models (BiGRU + attention). In the Explanation for the prediction task, various keyword-based matching techniques along with Rhetorical Roles (ratio) were used to find the court judgment decision.  Different span lengths from the ending portions of the legal judgments were considered for the explanation. 

%% file: results.tex
\section{Results and Discussion} \label{sec:results}

%\subsection{Baselines}
\noindent\textbf{Baselines:}
We used transformer-based models as a baseline for each of the tasks. The baseline model for RR tasks was SciBERT-HSLN \cite{brack2021sequential} (\url{https://github.com/Legal-NLP-EkStep/rhetorical-role-baseline}). We used RoBERTa base with the transition-based parser as the baseline for the NER task (\url{ https://github.com/Legal-NLP-EkStep/legal_NER}). Similarly, the baseline model for CJPE was a Hierarchical model with XL-Net at the first level and Bi-GRU at the next level { \url{https://github.com/Exploration-Lab/CJPE}}. 

%\subsection{Task A}
\noindent\textbf{Task A:}
Results for Task A are shown in Table \ref{table:Task-1-Results}. Overall, the use of BiLSTM layers with a CRF module seemed to be a popular, showing promising results. In general, the strategy of augmenting the dataset seemed to be a better choice which showed promising improvements and generalized predictions, and worked well on the unseen test set. One of the teams (\texttt{VTCC-NLP}) also used GNNs to capture the relationship between the entities to classify sentences into RR categories. The best-obtained results on the test set were around $86\%$ (Team \texttt{AntContentTech}) followed by scores ranging from $84\%$ to $80\%$ where the majority of the methods used pre-trained transformer-based architectures trained on Legal text (e.g., LegalBERT, InLegalBERT) as the primary backbone for legal text representations. The baseline model has a performance of 79\% F1 score. The proposed methods show an improvement over the baseline, however, there is further scope for improvement. 
%The best performing model is by the team \texttt{AntContentTech} who use LegalBERT+BiLSTM+CRF.   
%\AM{comparison with baseline results}

%\subsection{Task B}
\noindent\textbf{Task B:}
Results for Task B are shown in Table \ref{table:Task-2-Results}. Most of the teams used transformer-based backbone (e.g., RoBERTa, LegalBERT). Surprisingly, the use of the Legal-BERT-based performs similarly to standard pre-trained RoBERTa, highlighting the generic nature of named entity recognition tasks that may not require specialized legal knowledge. Augmentation techniques like back-translation seem to improve the performance of NER systems on Legal text. The best-performing system makes use of ensembles to further improve the results of the proposed system. In comparison to the baseline (F1 score of $91.1\%$), there was not much improvement in the best-performing system. This highlights the development of better techniques to improve the Legal-NER system further. 
%\AM{comparison with baseline results}

%\subsection{Task C}

\noindent\textbf{Task C:}
Table \ref{table:Task-3a-Results} and Table \ref{table:Task-3b-Results} show the results for the task of Legal Judgement Prediction (LJP) and Legal Judgment Prediction with Explanation (LJPE) respectively. For the LJP task, all the systems used  transformer-based models with the best-performing model using LongFormer which was able to capture long-range dependencies in legal documents. For the LJPE task, systems used occlusion-based and attention-based methods for the explanation. The best-performing system used Rhetorical Roles to extract the ratio of the case and that was subsequently used to provide the explanation, this is similar to the approach proposed by \citet{malik-etal-2022-semantic}. Viettel-AI \cite{hoang-bui-bui:2023:SemEval} group achieved the highest F1 score of $74.85\%$ in task C1 and Nonet \cite{nigam-EtAl:2023:SemEval} has the highest score of $54.18\%$ in task C2. 

%Baseline: F1 score of $76.55\%$

%% file: conclusion.tex
\section{Conclusion} \label{sec:conclusion}

In this paper, we present the findings of ‘SemEval 2023 Task 6: LegalEval - Understanding Legal Texts’, consisting of three Subtasks: i) Subtask A, Rhetorical Role identification, aimed to segment a legal document by assigning a rhetorical role to each of the sentence present in the document ii) Subtask B, Named Entity Recognition, for identifying NER tags present in a legal text and iii) Subtask C, Legal Judgment Prediction with Explanation, which aims to predict the final judgment of a legal case and in addition, provides explanations for the prediction. In total, 40 teams participated and 26 teams (approximately 100 participants) submitted systems papers and presented various models for each task. We provide an overview of methods proposed by the teams. 
Analysis of the systems and results points out more scope for improvement. We hope these findings will provide valuable insights to improve approaches for each of the sub-tasks.